\documentclass[10pt,twocolumn,letterpaper]{article}

\usepackage{iccv}
\usepackage{times}
\usepackage{epsfig}
\usepackage{graphicx}
\usepackage{amsmath}
\usepackage{amssymb}
\usepackage{xfrac}
\usepackage{booktabs}
\usepackage{multirow}
\usepackage[accsupp]{axessibility}  

\usepackage[pagebackref=true,breaklinks=true,letterpaper=true,colorlinks,bookmarks=false]{hyperref}
\usepackage{cleveref}

\iccvfinalcopy 

\def\iccvPaperID{8931} 

\begin{document}
\ificcvfinal\pagestyle{empty}\fi

\title{Latent Transformations via NeuralODEs\\ for GAN-based Image Editing}

\author{\textbf{Valentin Khrulkov}\textsuperscript{1}\thanks{Equal contribution} \quad \textbf{Leyla Mirvakhabova}\textsuperscript{2}\footnotemark[1] \quad \\
\textbf{Ivan Oseledets}\textsuperscript{2} \quad \textbf{Artem Babenko}\textsuperscript{1,3}\\
\\
Yandex\textsuperscript{1} \\
Skolkovo Institute of Science and Technology (Skoltech)\textsuperscript{2}\\
National Research University Higher School of Economics\textsuperscript{3}\\
{\tt\small khrulkov.v@gmail.com, \tt\small \{leyla.mirvakhabova,i.oseledets\}@skoltech.ru}, \tt\small artem.babenko@phystech.edu}

\maketitle

\begin{abstract}
   Recent advances in high-fidelity semantic image editing heavily rely on the presumably disentangled latent spaces of the state-of-the-art generative models, such as StyleGAN. Specifically, recent works show that it is possible to achieve decent controllability of attributes in face images via linear shifts along with latent directions. Several recent methods address the discovery of such directions, implicitly assuming that the state-of-the-art GANs learn the latent spaces with inherently linearly separable attribute distributions and semantic vector arithmetic properties. 
   
   In our work, we show that nonlinear latent code manipulations realized as flows of a trainable Neural ODE are beneficial for many practical non-face image domains with more complex non-textured factors of variation. In particular, we investigate a large number of datasets with known attributes and demonstrate that certain attribute manipulations are challenging to obtain with linear shifts only.
\end{abstract}

\section{Introduction}
Generative Adversarial Networks (GANs) \cite{goodfellow2014generative} have significantly advanced techniques for image processing and controllable generation, such as semantic image-to-image translation \cite{isola2017image, choi2018stargan, lee2020drit++, zhu2017unpaired, zhu2017toward} and image editing via manipulating the internal GAN activations \cite{bau2018gan, collins2020editing} or generator parameters \cite{bau2020rewriting, cherepkov2020navigating}. Moreover, since the GAN latent spaces are known to possess semantically meaningful vector space arithmetic, a plethora of recent works explore these spaces to discover the interpretable directions \cite{radford2015unsupervised, shen2020interpreting, goetschalckx2019ganalyze, jahanian2019steerability, plumerault2019Controlling, voynov2020unsupervised, harkonen2020ganspace, peebles2020hessian}. The directions identified by these methods are then used to manipulate user-specified image attributes, which is shown to be particularly successful for face images \cite{shen2020interpreting}.

While a large number of methods exploring the latent spaces of pretrained GANs have recently been developed, most of them learn linear latent controls, and more complex nonlinear latent transformations are hardly addressed. We conjecture that this limitation could arise because most of the latent editing literature is biased to the human face datasets, where linear transformations are sufficient for decent editing quality \cite{shen2020interpreting}.

In this work, we demonstrate that in the general case, the linear latent shifts cannot be used universally for all domains and attributes, and more complex nonlinear transformations are needed. To this end, we analyze how different attribute values are distributed in the latent spaces of GANs trained on several synthetic and real datasets with known attribute labels. Our analysis shows that for non-face images, many attributes cannot be controlled by linear shifts. To mitigate this issue, we propose an alternative parametrization of the latent transformation based on the recent Neural ODE work \cite{chen2018neural}. Our parametrization allows for gradient-based optimization and can be used within existing methods for latent space exploration \cite{shen2020interpreting}. Through extensive experiments, we show that the proposed nonlinear transformations are much more appealing for the purposes of controllable generation. In particular, we show that nonlinear transformations are more beneficial for edits requiring global content changes, such as changing appearance of a scene.

To sum up, our contributions are the following:
\begin{itemize}
    \item We analyze the distributions of different attribute values in the GAN latent spaces and show that linear latent controls are typically not sufficient beyond the human face domain.
    \item We propose a Neural ODE-based parametrization of the latent transformation that allows for learning the nonlinear controls. On several non-face datasets, we show that usage of this parametrization results in higher editing quality confirmed qualitatively and quantitatively.
    \item We propose a technique to analyze the learned Neural ODE models and reveal the attributes that require nonlinear latent transformations.
\end{itemize}

\section{Related work}
\textbf{Latent manipulations in GANs.} The prior literature has empirically shown that the GAN latent spaces are endowed with human-interpretable vector space arithmetic \cite{radford2015unsupervised, shen2020interpreting, goetschalckx2019ganalyze, jahanian2019steerability, voynov2020unsupervised, zhuang2021enjoy, spingarn2020gan}. E.g., in GANs trained on face images, their latent spaces possess linear directions corresponding to adding smiles, glasses, and gender swap \cite{radford2015unsupervised, shen2020interpreting}. Since such interpretable directions provide a straightforward route to powerful image editing, their discovery currently receives much research attention. A line of recent works \cite{goetschalckx2019ganalyze, shen2020interpreting} employs explicit human-provided supervision to identify interpretable directions in the latent space. For instance, \cite{shen2020interpreting} use the classifiers pretrained on the CelebA dataset \cite{liu2015deep} to predict certain face attributes. These classifiers are then used to produce pseudo-labels for the generated images and their latent codes. Based on these pseudo-labels, the separating hyperplane is constructed in the latent space, and a normal to this hyperplane becomes a direction, controlling the corresponding attribute. Another work \cite{goetschalckx2019ganalyze} solves the optimization problem in the latent space, maximizing the score of the pretrained model, which predicts the image aesthetic appeal. The result of this optimization is the direction that makes images more aesthetically pleasing. Two self-supervised works \cite{jahanian2019steerability, plumerault2019Controlling} seek the vectors in the latent space that correspond to simple image augmentations such as zooming or translation. Finally, a bunch of recent methods \cite{voynov2020unsupervised, harkonen2020ganspace, peebles2020hessian} identify interpretable directions without any form of (self-)supervision. \cite{voynov2020unsupervised} learns a set of directions that can be easily distinguished by a separate classification model based on two samples, produced from the original latent codes and shifted along the particular direction. \cite{peebles2020hessian} learns the directions by minimizing the sum of squared off-diagonal terms of the generator Hessian matrix. Another approach, \cite{harkonen2020ganspace}, 
demonstrates that interpretable directions often correspond to the principal components of the activations from hidden layers of generator networks.

\textbf{Nonlinear editing.} While some works \cite{jahanian2019steerability, abdal2020styleflow, zhuang2021enjoy} briefly mention the possibility of non-linear latent transformations, they do not provide reliable evidence of the necessity of non-linear editing; therefore, most of the recent editing literature employs only linear manipulations. To the best of our knowledge, our work is the first that demonstrates several cases of inadequacy caused by linear editing and provides a rigorous quantitative comparison with non-linear techniques on several datasets.

\section{GAN-based image editing}
In this section, we remind on current approaches to controllable image generation and editing via GANs and discuss their possible weaknesses.

We assume that we are given a well-trained GAN model $G: \mathcal{W} \to \mathcal{X}$, where $\mathcal{W} \subset \mathbb{R}^d$ denotes the latent space and $\mathcal{X} \subset \mathbb{R}^{C \times H \times W}$ is the image space. We work with the style--based generators where the manipulation is performed in the so-called \emph{style space} $\mathcal{W}$, which has been shown to be more ``disentangled'' with respect to various image attributes.
We focus on the supervised setting and assume that we are given a trained semantic attribute regressor network $\mathcal{R}: \mathcal{X} \to \mathcal{S} \subset \mathbb{R}^N$, which predicts the attribute values for a given image. Here $\mathcal{S}$ denotes the semantic attribute space of the image domain $\mathcal{X}$, \eg, for human faces, this can be hair color, age, \etc. The space of image attributes $\mathcal{S}$ may be exhaustive, \ie, a point $\mathbf{x} \in \mathcal{X}$ may be uniquely determined by its attributes $\mathcal{R}(\mathbf{x})$, or be only a subset of a ``true'' attribute space.

\subsection{Manipulating image attributes by shifts}
Most of the current approaches propose to manipulate attributes of synthesized images with simple \emph{linear} translations in the latent space. This means the following. Let $\mathbf{s}$ denote the attribute vector of a generated image $G(\mathbf{w}_{0})$, and let $\mathbf{s}_i$ be a single chosen (binary) attribute. These approaches seek to carefully construct a vector $\mathbf{n}_i$ such that by gradually changing $\mathbf{w}_{0}$ as 
\begin{equation}\label{eq:linear}
    \mathbf{w}(\alpha) = \mathbf{w}_{0} + \alpha \mathbf{n}_i,
\end{equation}
we achieve controls over the value of the attribute $\mathbf{s}_i$. Note that these approaches assume that we use a \emph{single} vector $\mathbf{n}_i$ for all the points $\mathbf{w} \in \mathcal{W}$. InterFaceGAN \cite{shen2020interpreting} is among the most successful approaches to construct the shift vector that manipulates a desired attribute in the supervised setting. The idea of this method is to find a \emph{hyperplane} in the latent space separating $\mathbf{w}$ with different values of $\mathbf{s}_i$. For a large number of random style vectors $\mathbf{w}$ the labels are obtained by evaluating $\mathcal{R}[G(\mathbf{w})]$, and the hyperplane is found by fitting an SVM on this synthetic labeled dataset. The corresponding direction $\mathbf{n}_i$ is then simply a normal vector to this hyperplane.

\begin{figure}
    \centering
    \includegraphics[width=0.6\linewidth]{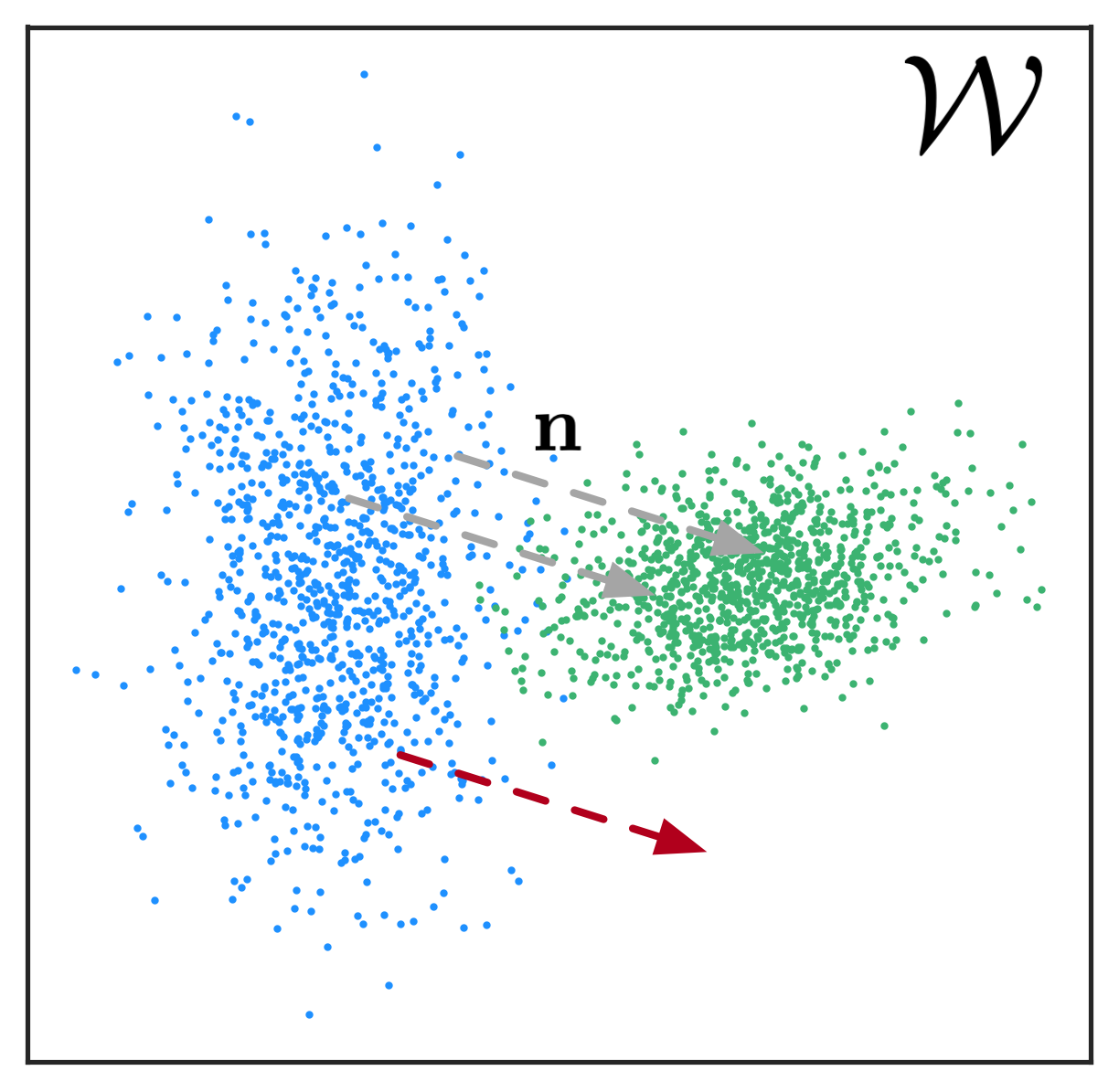}
    \caption{Consider the following toy example of a distribution of a binary attribute $\mathbf{s}_i$ in the latent space $\mathcal{W}$. With any translation vector $\mathbf{n}$, certain points in the left distribution will ``miss'' the right distribution. This suggests that more complex, \ie, nonlinear translations may be necessary.}
    \label{fig:example}
    \vspace{-0.5cm}
\end{figure}

\section{Nonlinear approach\label{sec:nonlinear}}
In contrast to previously described methods, in this section, we focus on the nonlinear approach to manipulations in GAN latent spaces. We can view the simple linear shift given by~\eqref{eq:linear} as the flow of a differential equation with a constant righthand side, \ie, $\dot{\mathbf{w}} = \mathbf{n}_i$ with the initial condition $\mathbf{w}(0)=\mathbf{w}_0$. The generalization of such edits to the nonlinear domain can be made straightforwardly: \eg, by replacing the righthand side with some function, depending on the input $\mathbf{w}$. We propose a simple approach: we consider the Neural ODE model \cite{chen2018neural} with a righthand side parametrized by a neural network consisting of a few linear layers with the Leaky ReLU activation. The model is later trained end-to-end using the regressor $\mathcal{R}$. Let us now discuss the model structure and training procedure in detail.

\subsection{Reminder on Neural ODEs}

The Neural ODE model \cite{chen2018neural} bridges the differential equations and neural networks by parametrizing a system of ODEs:
\begin{equation}{\label{eq:ode}}
    \dot{\mathbf{h}}(t) = f(\mathbf{h}(t), t; \mathbf{\theta}),
\end{equation}
 where $t \in [0, T]$ is time and $\mathbf{h}(t) \in \mathbb{R}^d$.
The solution of the ODE problem at time step $t=T$ serves as the output of the corresponding hidden layer, where the input is provided as the initial value to~\eqref{eq:ode}. In practice, the output can be computed via black-box differential equation solvers. In order to compute gradients with respect to $\theta$, it is common to use the adjoint method, which allows for memory-efficient backpropagation at the cost of extra function evaluations.

\subsection{Neural ODE for image manipulation.}
We directly apply Neural ODEs for image manipulation performed in the latent space of GANs. \Ie, we replace the linear flow \eqref{eq:linear} with the curved flow of a trainable Neural ODE in the latent space. Let us now briefly describe the specifics and the optimization goal.

\paragraph*{Network architecture.}
The righthand side of Neural ODE model $f(\cdot; \theta)$ is represented by a simple multilayer perceptron (MLP) with Leaky ReLU nonlinearity (with $\alpha=0.2$). We vary the number of layers in the network $f$ from $1$ to $3$; we additionally consider a constant righthand side, \ie, equation of the form $\dot{\mathbf{w}} = \theta$ with trainable $\theta$. We additionally normalize the righthand side of the ODE to the unit length, so the trajectories for all the approaches have the same length for the same value of $T$. To sum up, our Neural ODE takes the following form.
\begin{equation}
    \dot{\mathbf{w}} = \frac{f(\mathbf{w}; \theta)}{\|f(\mathbf{w}; \theta)\|},
\end{equation}
with $f(\cdot; \theta)$ being an MLP (or constant) as described above. To compute image edits, we then move along the trajectories of this ODE in the latent space.

\paragraph*{Loss function.}
As was discussed above, we search for transformations in the latent space in such a manner that they would change the desired attribute while leaving the others unchanged. Recall, that $\mathcal{R}$ is a network, which predicts the value of image attributes. Suppose that we have $N$ discrete attributes and our goal is to manipulate the $i$-th (binary) attribute. Let $\mathbf{w}$ be a random style vector with a vector of attributes $\mathcal{R}(\mathbf{w}) = (\mathbf{s}_1, \hdots, \mathbf{s}_i, \hdots \mathbf{s}_N)$. We set the target attribute vector $\hat{\mathbf{s}} = (\mathbf{s}_1, \hdots, 1 - \mathbf{s}_i, \hdots \mathbf{s}_N)$. After following along the trajectory of the Neural ODE starting at $\mathbf{w}$ for some time value $T$, we obtain a point $\mathbf{w}(T, \theta)$. More concretely, in practice we set the maximal value $T_{\mathrm{max}}$ of order $8-12$ and then randomly sample the interval $[\sfrac{T_{\mathrm{max}}}{4}, T_{\mathrm{max}}]$ to get the final time step (as was done in \cite{voynov2020unsupervised}). In what follows, as a slight abuse of notation, we will denote by $\mathcal{R}[\cdot]$ the predicted attribute values of a generated image: $\mathcal{R}[G(\cdot)]$.

To achieve the aforementioned desired transformation properties, we introduce a loss function consisting of two terms: the first one, denoted by $\mathcal{L}_1$, measures the discrepancy between the obtained and the desired $i$-th attribute values.
\begin{equation}{\label{eq:l1}}
\mathcal{L}_1(\mathbf{w}, \theta) = \mathrm{CE}(\mathcal{R}[\mathbf{w}(T, \theta)]_i, \hat{\mathbf{s}}_i),
\end{equation}
where $\mathrm{CE}$ stands for the cross entropy loss.
The second one represented by $\mathcal{L}_2$ is a loss term controlling the change of remaining attribute values. 
\begin{equation}{\label{eq:l2}}
\mathcal{L}_2(\mathbf{w}, \theta) = \frac{1}{N-1}\sum_{j=1, i\neq j}^{N}\mathrm{CE}(\mathcal{R}[\mathbf{w}(T, \theta)]_{j}, \hat{\mathbf{s}}_j).
\end{equation}
Finally, the loss function takes the form $\mathcal{L} = \mathcal{L}_1 + \mathcal{L}_2$.
 Note that this loss function, in general, can not be written as a single cross-entropy loss since the discrete attributes $\hat{\mathbf{s}}_j$ may belong to spaces of different cardinalities (e.g., we may have an attribute like `object position' assuming a large set of intermediate values). In our work, we search for a separate Neural ODE for each attribute; however, in principle, it is possible to consider conditional Neural ODEs and have a single model. 

\subsection{Evaluation}
Estimation of the quality of the image editing approach is a nontrivial task, often relying on mean opinion scores provided by human evaluators or some proxy metrics. Typically, given a method to manipulate the latent code of an image, we continuously steer it while visually observing whether the desired attribute shift happened and how disentangled the transformation looks \cite{shen2020interpreting,voynov2020unsupervised}. In a nutshell, we propose to measure these effects numerically in the spirit of traditional PR or ROC curves. Concretely, given a starting point $\mathbf{w}$ and a timestep $\tau$, we measure (i) whether the value of the attribute $\mathbf{s}_i$ is equal to the desired target value and (ii) for each of the remaining attributes we compute the normalized \emph{entropy} of the label distribution along the trajectory up to $\tau$. The idea behind (ii) is that in the ideal case, the attribute values remain constant, and relatively ``rare'' and localized spontaneous attribute changes are still satisfactory. Formally, for a given $\mathbf{w}$ with the attribute vector $\mathcal{R}(\mathbf{w}) = (\mathbf{s}_1, \hdots, \mathbf{s}_i, \hdots \mathbf{s}_N)$ and the target attribute vector $\hat{\mathbf{s}} = (\mathbf{s}_1, \hdots, 1 - \mathbf{s}_i, \hdots \mathbf{s}_N)$ these two metrics, termed $\mathrm{C}(\tau, \mathbf{w})$ and $\mathrm{D}(\tau, \mathbf{w})$ for \textbf{C}ontrol and \textbf{D}isentanglement, are defined as follows.
\begin{equation}{\label{eq:c_def}}
    \mathrm{C}(\tau, \mathbf{w}) = 
    \begin{cases}
    1, \hat{\mathbf{s}}_i = \mathcal{R}[\mathbf{w}(\tau)]_i \\
    0, \mathrm{otherwise}
    \end{cases},
\end{equation}

\begin{equation}{\label{eq:d_def}}
    \mathrm{D}(\tau, \mathbf{w}) = 
    \frac{1}{N-1} \sum_{j=1, j\neq i}^{N}\frac{\mathcal{H}\Big(\{\mathcal{R}[\mathbf{w}(t)]_j\}_{t=0}^{\tau}\Big)}{\mathcal{H}\Big(\mathrm{Uniform}(\#\mathcal{S}_j)\Big)}.
\end{equation}
Here $\#\mathcal{S}_j$ denotes the cardinality of $j$-th attribute, and $\mathcal{H}$ is the entropy. To get the global values of these metrics, we simply average them across a large number of samples. I.e, we obtain a curve $(\mathrm{C}(\tau) = \sfrac{1}{N} \sum_i C(\tau, \mathbf{w}_i), \mathrm{D}(\tau) = \sfrac{1}{N} \sum_i D(\tau, \mathbf{w}_i))$, which by construction lies in the unit square. By comparing relative positions of these curves for two methods, we can judge which provides more disentanglement/better control quality.
Note that, however, the reliable estimation of the quality by these metrics is only possible in the case when all the factors of variation in data are known, which is possible for synthetic datasets. For large-scale datasets of real images, we have to resort to standard visual evaluation with human assessors.

\section{Experiments\label{sec:experiments}}
\begin{figure*}[htb!]
    \centering
    \includegraphics[width=0.8\linewidth]{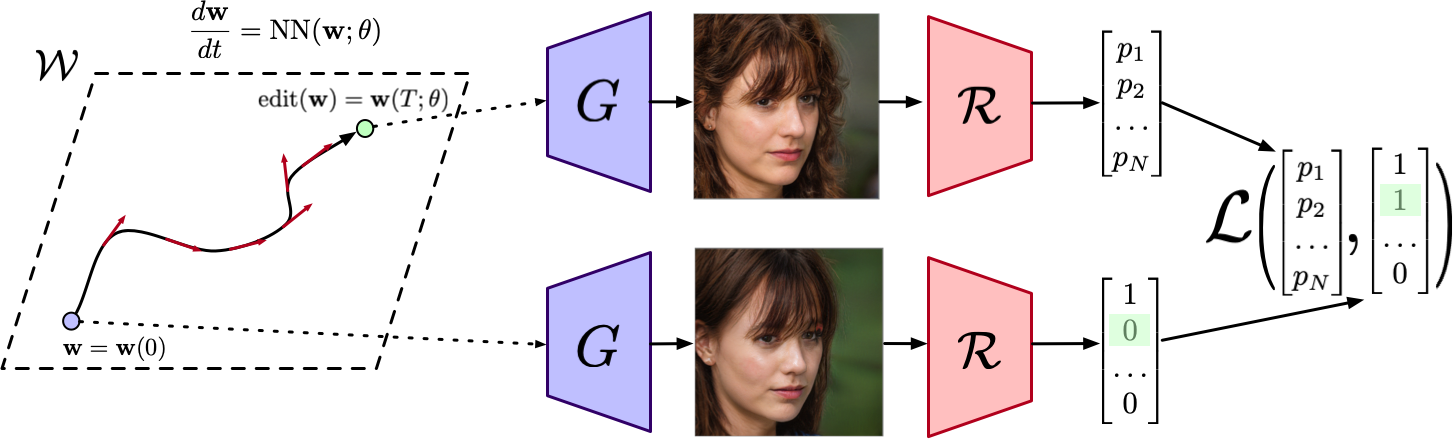}
    \caption{Our learning method. A sample from the latent space is transformed via the nonlinear flow of a trainable Neural ODE. The loss function ensures that the desired semantic attribute of the edited image changed while others remained fixed. The attributes are obtained using the pre-trained attribute regressor $\mathcal{R}$.}
    \label{fig:level_sets}
\end{figure*}

We have implemented the proposed method in \texttt{Pytorch}. For GAN training, we utilized a single DGX-1 station with $8$ Nvidia Tesla V100 GPUs, and for training Neural ODEs, we used a single V100 GPU (in our setting, it takes roughly $30$ minutes per single direction). Specifics of architectures, optimization details, and additional experiments, are available in the supplementary material. Our code and models are available at \href{https://github.com/KhrulkovV/nonlinear-image-editing}{github}.
\subsection{Synthetic datasets}
\begin{figure}[htb!]
    \centering
    \includegraphics[width=0.9\linewidth]{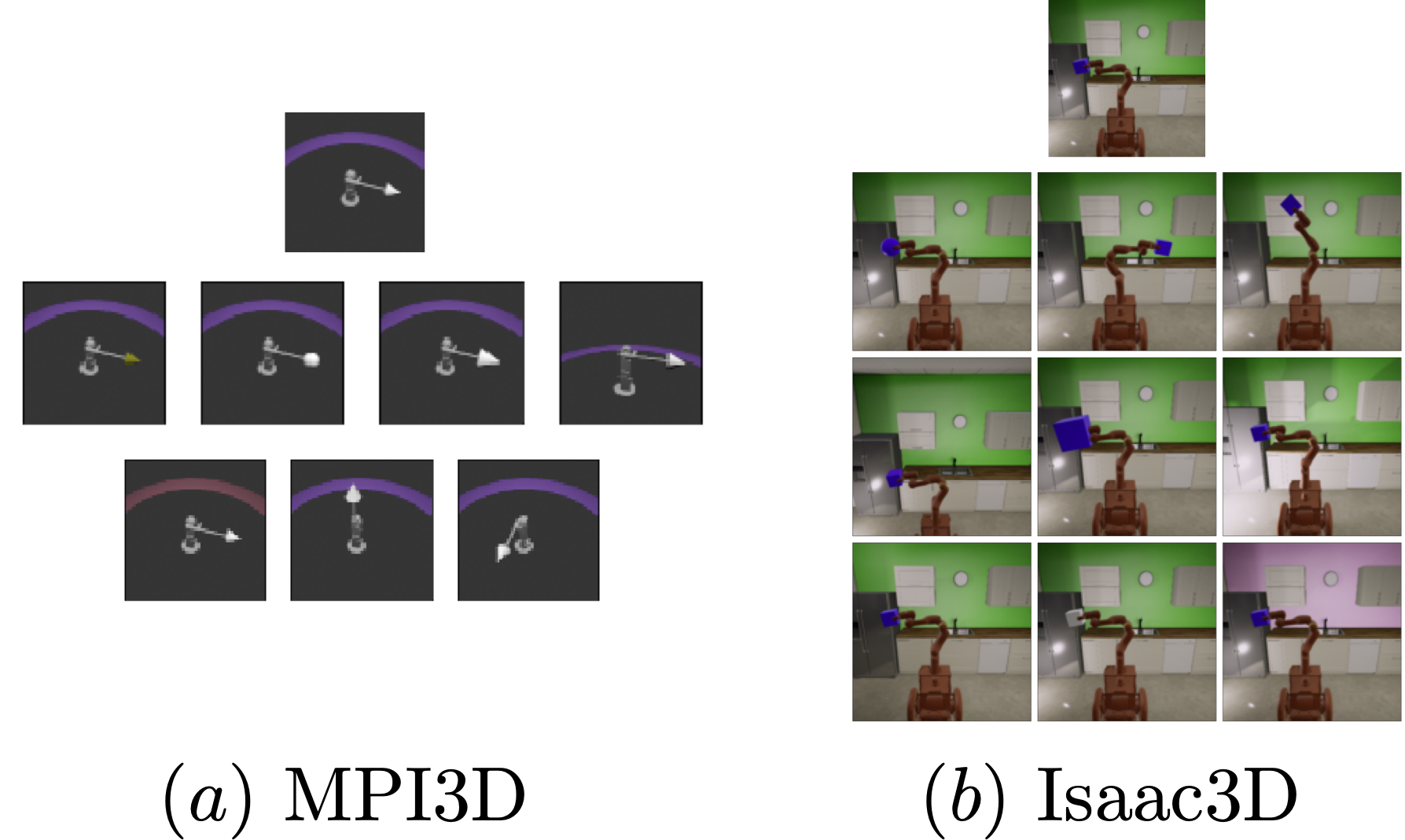}
    \caption{Samples from two synthetic datasets used for quantitative evaluation of the methods. In both cases, all the factors of variation are known.}
    \label{fig:dataset_example}
    \vspace{-0.5cm}
\end{figure}

The goal of this section is to quantitatively verify the benefits of nonlinear editing on several large-scale datasets where the ground truth factors of variations are known and contain both texture and non-texture-based attributes (see \Cref{fig:dataset_example} for an example).

\paragraph{Datasets.}
\begin{itemize}
    \item \emph{MPI3D} consisting of $1,036,800$ images with $7$ factors \cite{gondal2019transfer}. This dataset represents a robotic arm in various positions holding an object of varied shapes and colors. We use the toy part of the dataset, \ie, simply rendered images.
    \item \emph{Isaac3D} is a recently proposed dataset of high-resolution images \cite{nie2020semi} containing $737,280$ images with $9$ factors of variations; we resize images to $128 \times 128$ resolution. This is, in a way, an advanced version of \emph{MPI3D} with photorealistic images and more attributes.
\end{itemize}
For both of these datasets, each image is \emph{uniquely determined} by the corresponding attributes; thus, we can reasonably compare linear and nonlinear manipulations using the aforementioned metric.

\paragraph{GAN Model.}
We use the recently proposed StyleGAN 2 \cite{karras2020analyzing} and its implementation in \texttt{Pytorch} found at \href{https://github.com/rosinality/stylegan2-pytorch}{github}. We use the default settings except for the number of layers in the style network, which we set to $3$, as was done in \cite{nie2020semi,khrulkov2021disentangled}. For \emph{MPI3D} we trained the model for $12.5\mathrm{M}$ frames and for \emph{Isaac3D} for $25\mathrm{M}$ frames. We do not use any data augmentation for training. 
\paragraph{Attribute regressors.}
For each dataset, we train an attribute regressor network on the real data. For \emph{MPI3D}, we use a simple four-block CNN as a backbone, followed by multiple classification heads, and for \emph{Isaac3D} we use ResNet18 (not pretrained) as a backbone. In both cases, the attribute regressors were able to achieve more than $99\%$ accuracy on all the attributes on the test set.
\paragraph{Neural ODE models.}
We consider two Neural ODEs, namely one with a trainable constant righthand side (termed \textbf{Ours(linear)} on the plots), and righthand sides represented by MLPs of depth one (\textbf{Ours(nonlinear)}). We use the open-source implementation of Neural ODE found at \href{https://github.com/rtqichen/torchdiffeq}{github}. We train all the models for $5000$ iterations with a batch size of $24$. For these two datasets, all the attributes take more than two values, while previously, we considered binary attributes. To alleviate this, when rectifying the attribute with index $j$ for the sake of simplicity, we binarize it by learning to transform $\mathbf{s}_j=0$ to $\mathbf{s}_j=\#\mathcal{S}_j - 1$, and all other attributes retain their full discrete set of values. For both datasets we used $T_{\mathrm{max}}=12$. For \emph{Isaac3D}, we consider all the attributes, and for \emph{MPI3D}, we train Neural ODE models for the first five attributes due to the large cardinality of the two positional attributes; however, we still include them when computing metrics and during training.
As a reference, we include the scores obtained by the InterFaceGAN (\textbf{IF}) method and its `disentangled' version, termed \textbf{IF projected}. The latter was obtained using the conditional manipulation approach specified in \cite{shen2020interpreting} (we conditioned each attribute on all the remaining attributes).

\subsubsection{Evaluation of the learned manipulations}
\begin{figure}
    \centering
\includegraphics[width=0.85\linewidth]{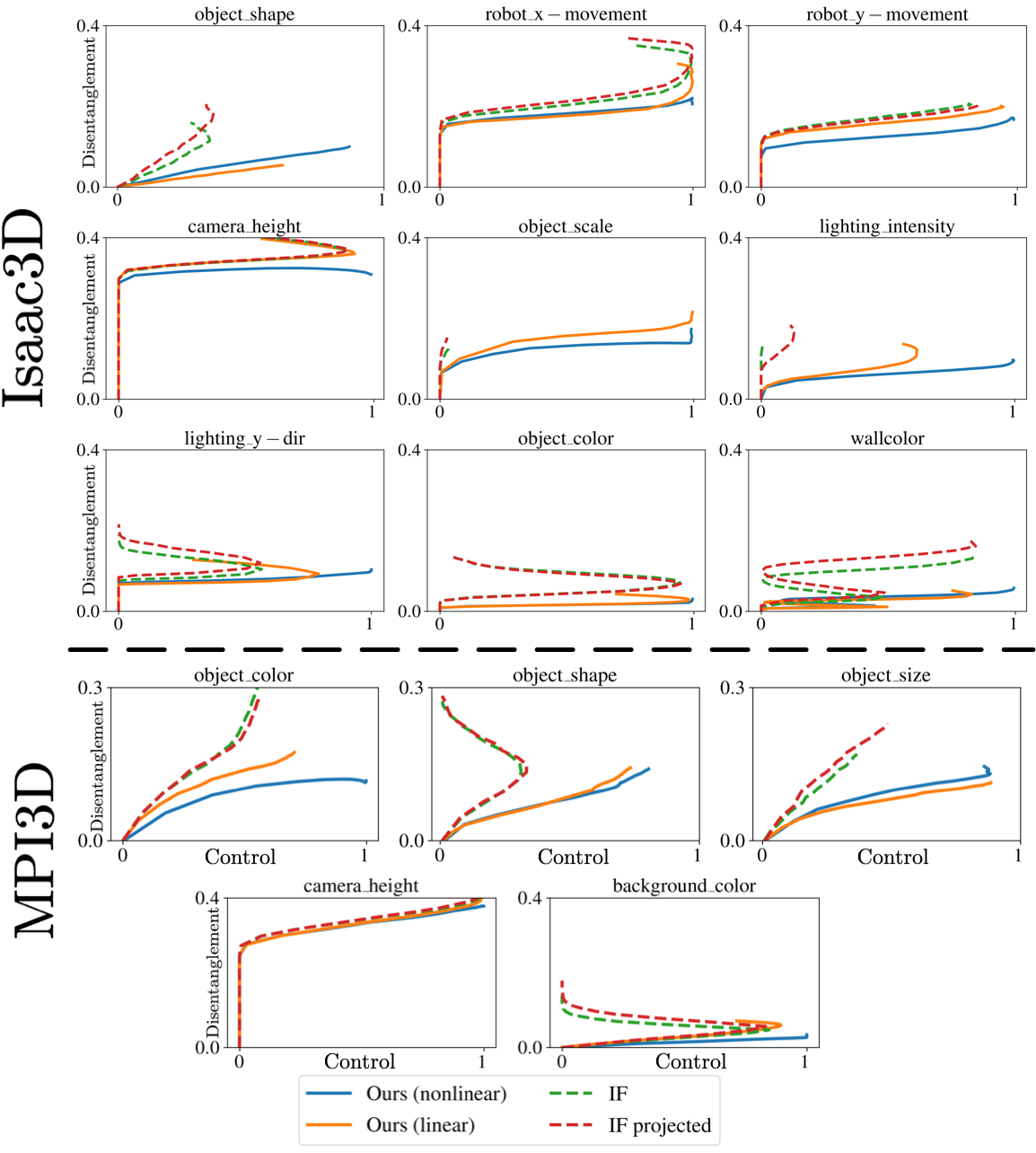}
    \caption{Control-Disentanglement curves for \emph{Isaac3D} and \emph{MPI3D}. We observe that, unlike linear shifts, nonlinear flows allow for achieving good control for all the samples while maintaining reasonable disentanglement.}
    \label{fig:merged_cd}
    \vspace{-0.5cm}
\end{figure}
\begin{figure}
    \centering
    \includegraphics[width=0.8\linewidth]{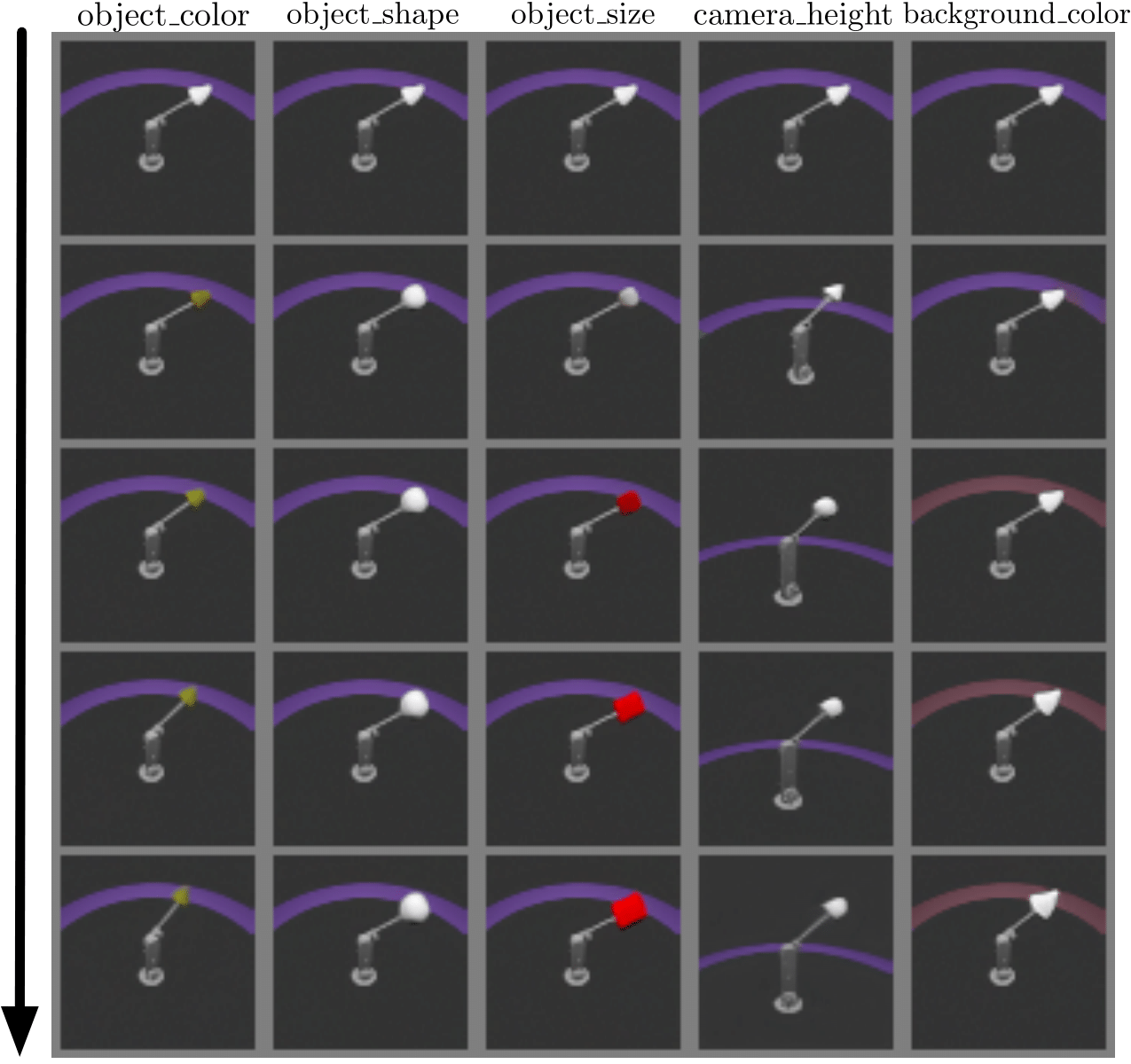}
    \caption{Manipulating the first five attributes on \emph{MPI3D}. For the visualization, we used learned Neural ODEs of depth $2$.}
    \label{fig:mpi_example}
    \vspace{-0.5cm}
\end{figure}

We now evaluate the obtained Neural ODEs. Results for \emph{Isaac3D} are summarized on \Cref{fig:merged_cd}. Here we plot \mbox{CD-curves} as was discussed in \Cref{sec:nonlinear}. Intuitively, a lower position of one curve, well-covering the Control range, with respect to another, indicates better quality of the image editing method. We observe that deep Neural ODEs can obtain a reasonable trade-off between disentanglement and control for all the attributes. On the other hand, linear controls are inferior in terms of either control (i.e., they do not work for a subset of the latent space) or provide inferior disentanglement. Interestingly, in some cases, the curves make a jump near the origin, \eg, for \texttt{camera\_height}. Such behavior indicates that the latent codes have to travel a considerable distance before the attribute shift occurs, which intuitively corresponds to neat well--separated attribute distributions. On the other hand, for many other non-textured attributes, such distributions may ``interlace'' the latent space $\mathcal{W}$ and the attribute transition can occur relatively close to the point of origin. An example of learned manipulations is provided at \Cref{fig:mpi_example}.

\subsection{Real-life datasets}
In this section, we investigate the behavior of nonlinear image edits learned by our Neural ODE-based approach on real-life datasets. Additionally, we include experiments on the \emph{CUB-200-2011} dataset in \Cref{app:cub}.
\paragraph*{Datasets.} 
\begin{itemize}
    \item \emph{FFHQ} is a dataset consisting of $70,000$ high-quality human faces images \cite{karras2019style}. This is a standard benchmark for image editing as it contains a rich variation in age, ethnicity, lighting, and background.
    \item \emph{Places365} consists of $1,803,460$ training images with $400+$ unique scene categories \cite{zhou2017places}. We restrict our dataset to the \texttt{outdoor natural} scenes and filter out the ones with the attribute \texttt{man-made}. The final version contains $48$ classes and $239,457$ images.
\end{itemize}
\paragraph*{Model.} For these datasets, we also used StyleGAN 2. For \emph{FFHQ}, we used the recent high--quality pretrained model producing images of resolution $256 \times 256$ and provided by the authors of \cite{karras2020training} at \href{https://github.com/NVlabs/stylegan2-ada}{github}. \emph{CUB-200-2011} and \emph{Places365} are especially challenging datasets for generative modeling due to the low number of samples and high sample diversity. We utilize the Adaptive Data Augmentation (ADA) strategy \cite{karras2020training}, which helps to deal with limited size datasets. We use the authors' implementation in \texttt{tensorflow} at the same github link. We train both models for $25\mathrm{M}$ frames with the default config; we only change the number of layers in the style network to $8$ to be consistent with the \emph{FFHQ} model. For training, we resize all the images to $256 \times 256$.
\paragraph*{Attribute regressors.}
For all the attribute regressors, we use the same frozen ResNet18 backbone pretrained on ImageNet, followed by a trainable MLP of depth two, where for each attribute, we consider a separate classification head. To train the regressors, we utilized the following data.
\begin{itemize}
    \item For \emph{FFHQ}, we used the data and attribute annotation provided by the CelebA \cite{liu2015deep} dataset. There are $202,599$ images with $40$ binary attributes such as smile, gender, hairstyle, \etc.
    \item For \emph{Places365} we used the Transient Attributes \cite{Laffont14} dataset. This dataset contains $8,571$ scene images with annotations for $40$ binary attributes such as ``fog'', ``snow'', ``dusk'', ``autumn''. We used all $40$ attributes when training the regressor.
\end{itemize}
Another approach to enforce identity preservation in our method is to utilize an off-the-shelf representation network $\mathcal{F}$, such as FaceNet \cite{schroff2015facenet} for human faces datasets. In this case, we replace our $\mathcal{L}_2$ loss with the cosine distance between the representations $\mathcal{F}[G(\mathbf{w}(T; \theta)]$ and $\mathcal{F}[G(\mathbf{w})]$. See supplementary material for the details on this experiment.

\paragraph*{Neural ODE.}
We use exactly the same settings and loss function as for the synthetic datasets. We train a separate model for each attribute. We experimented with Neural ODEs of depth $1$ and $2$, but we did not notice any significant visual difference, so we chose to stick with depth $1$ in our visualizations. We denote it by $\mathbf{Ours(nonlinear)}$. To verify the actual benefit of \emph{nonlinearity} over simply having a more powerful loss function, we also study Neural ODEs with a trainable constant righthand side. We denote it by $\mathbf{Ours(linear)}$.
\paragraph*{Baselines.}
As the baseline approach for supervised image editing, we consider InterFaceGAN (IF). We use $20,000$ latent codes to train SVMs. We did not obtain competitive results with the conditional IF, thus we utilize the standard version. This approach is similar to other works \cite{abdal2020styleflow,zhuang2021enjoy}.

\subsubsection{FFHQ}
\begin{figure}
    \centering
\includegraphics[width=0.9\linewidth]{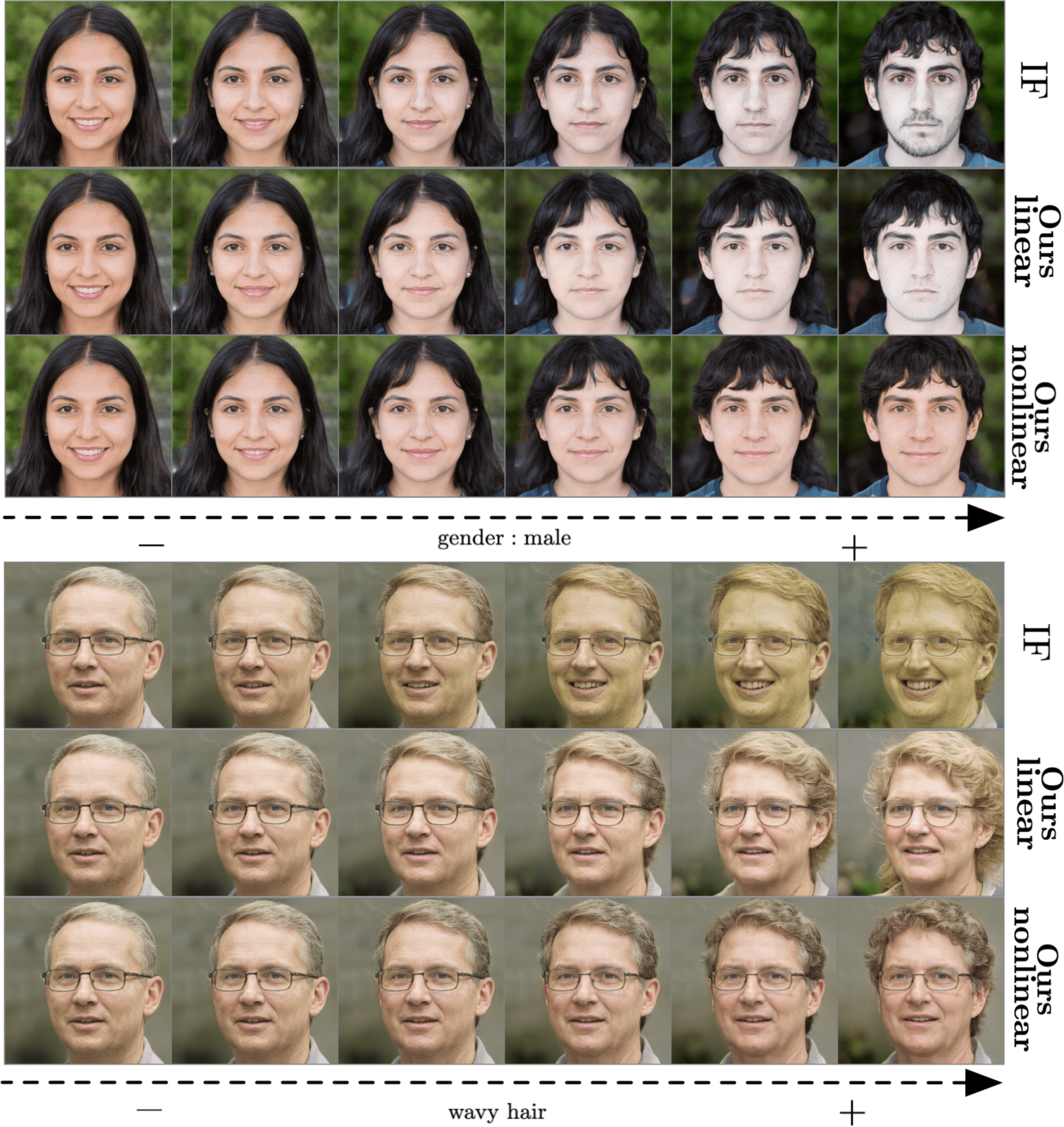}
    \caption{Manipulating two different attributes on the StyleGAN 2 trained on the \emph{FFHQ} dataset. For \texttt{gender:male} and \texttt{wavy hair} linear shifts suffer from (i) unnatural face color (ii) identity change.}
    \label{fig:celeba}
\end{figure}

\begin{figure}
    \centering
\includegraphics[width=0.9\linewidth]{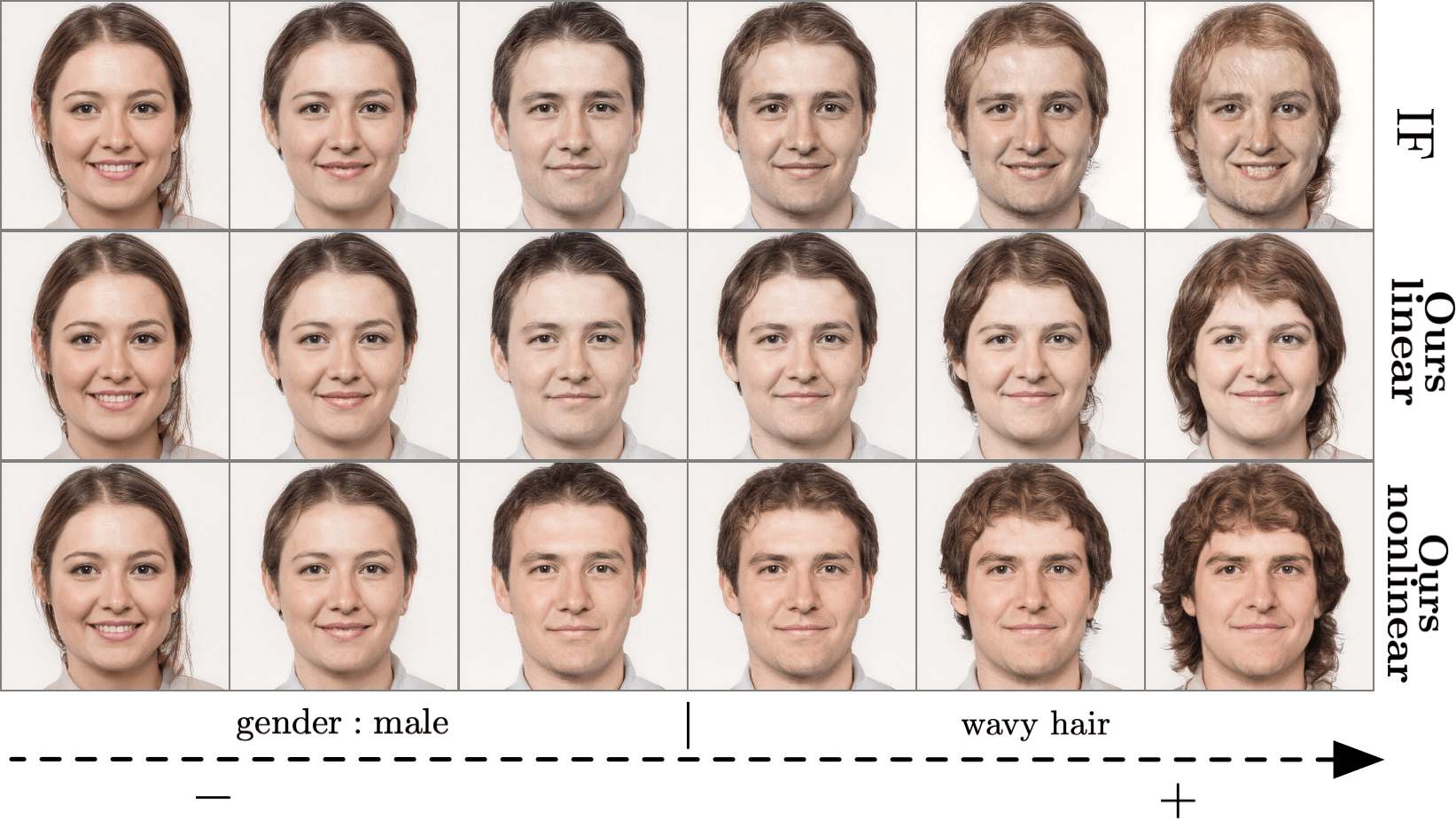}
    \caption{An example of a sequential attribute manipulation: \texttt{gender:male} combined with the subsequent \texttt{wavy hair} attribute. Our nonlinear method performs visually better with respect to both control and disentanglement.}
    \label{fig:celeba_composition}
    \vspace{-0.4cm}
\end{figure}

We hypothesize that for datasets consisting of human face images, the attributes describing \emph{texture}-based features (\eg, hair or skin color) can be manipulated linearly relatively well, while for \emph{non-texture}-based attributes (\eg, hair type, gender), the linear shifts may have slightly worse performance. To support our hypothesis, we experiment with \texttt{gender} and \texttt{wavy hair}; our findings are described in \Cref{fig:celeba}. Additionally, we experiment with a \emph{composition} of attribute manipulations: \eg, we may want to change the gender at first and then manipulate the hair type; our experiments are summarized in \Cref{fig:celeba_composition}. We note that in all the experiments, our nonlinear method outperforms or is on par with linear methods in terms of visual quality.

\subsubsection{Places365}
\begin{figure}[htb!]
    \centering
\includegraphics[width=0.9\linewidth]{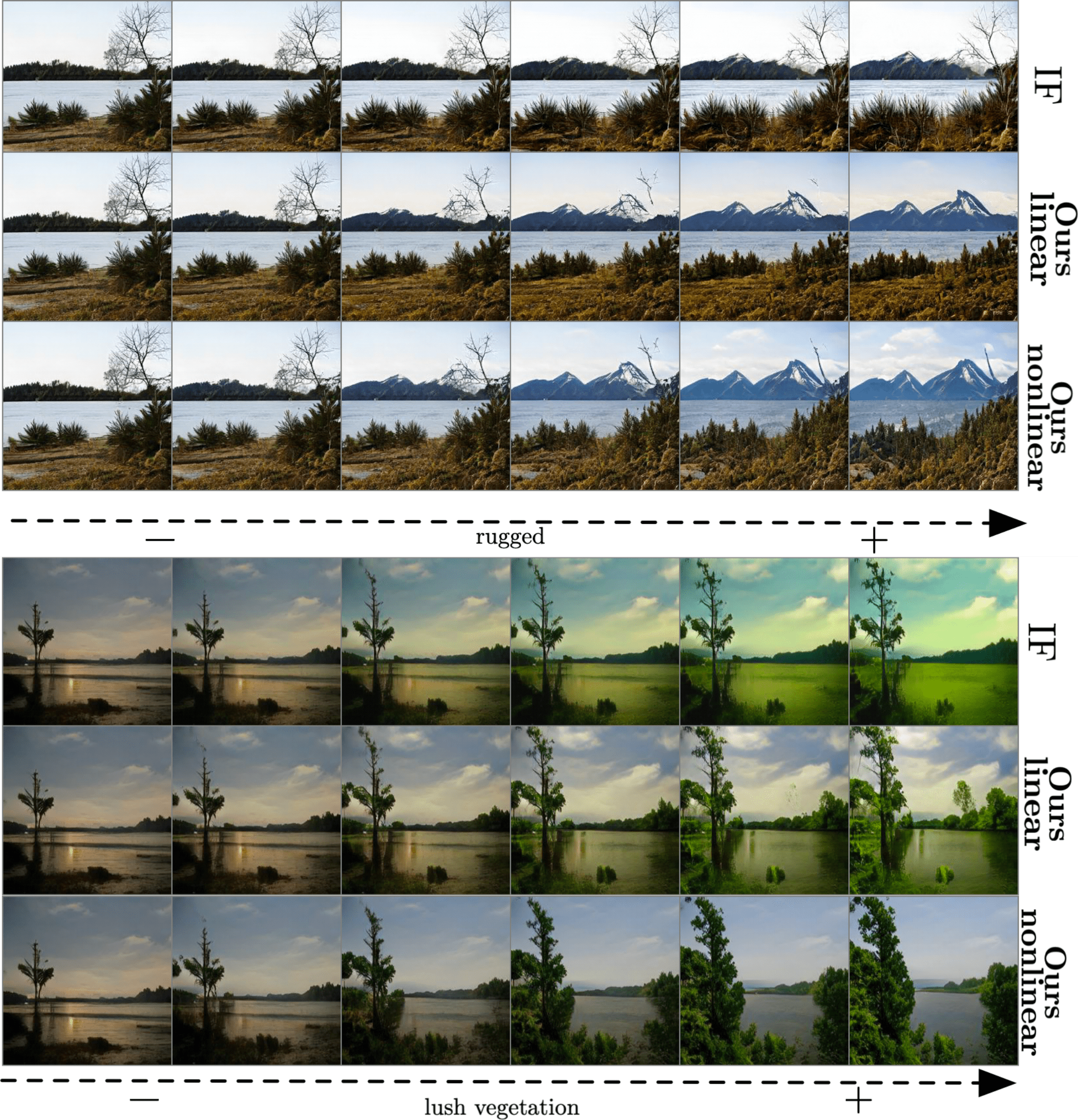}
    \caption{Manipulating \texttt{rugged} and \texttt{lush vegetation} attributes on the \emph{Places365} dataset. We observe that our nonlinear method achieves the desired control over image contents, while linear shifts tend to change images' texture.}
    \label{fig:all_scenes}
    \vspace{-0.3cm}
\end{figure}

Similar to the previous reasoning, we consider the attributes which intuitively correspond to the drastic change of image content. Namely, we study the \texttt{rugged} attribute and the \texttt{lush vegetation} attribute. Results are provided at \Cref{fig:all_scenes}. Here we can observe an interesting failure mode of linear methods: for instance, in the last example, they simply make the texture greener, which on a very high level, corresponds to more ``vegetation''. However, they struggle to add details like trees or grass, which is successfully achieved by our nonlinear method. Similar results hold for the \texttt{rugged} attribute as well.

\subsubsection{Editing real images}
\begin{figure}[htb!]
    \centering
\includegraphics[width=0.9\linewidth]{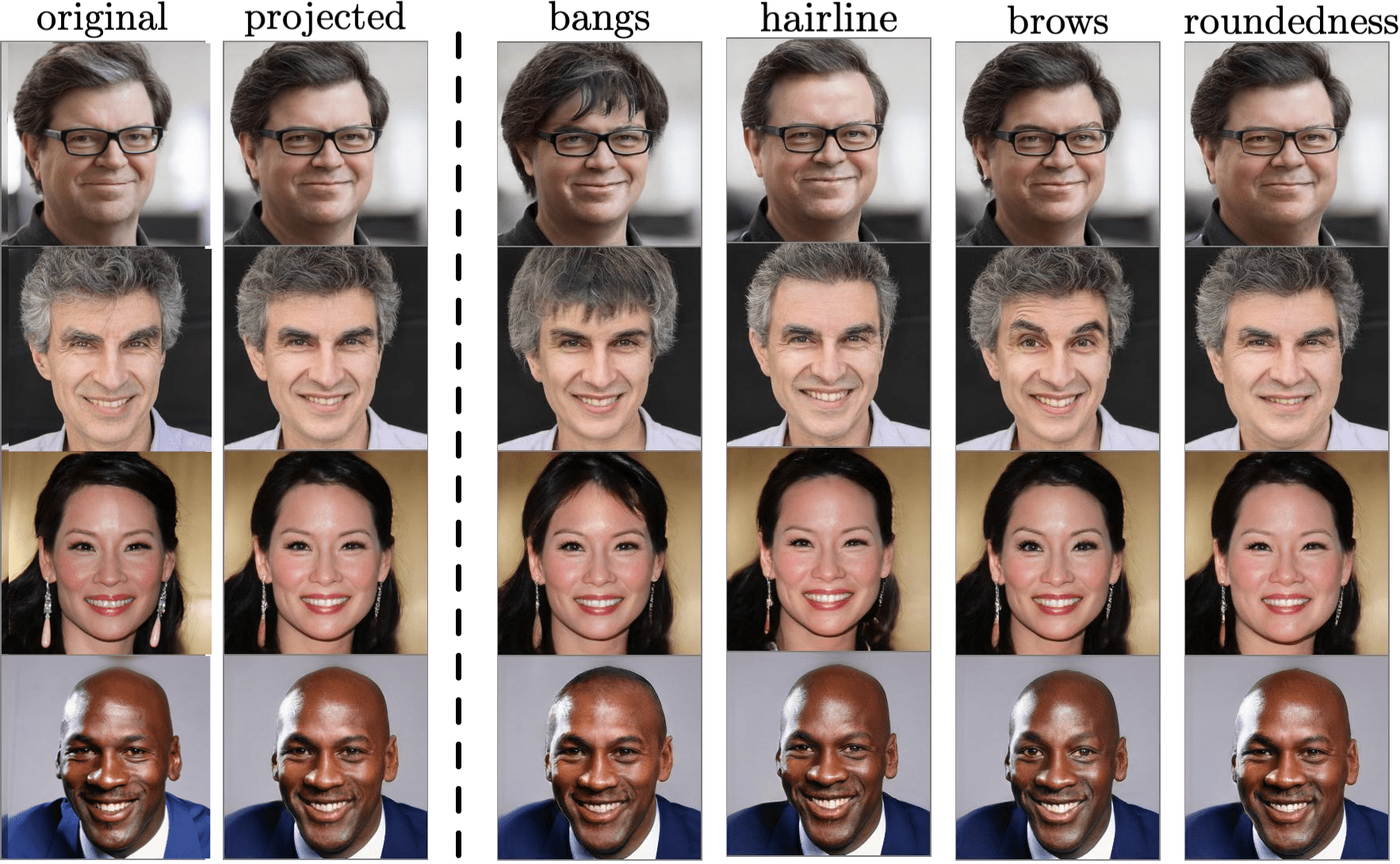}
    \caption{Manipulation of real images embedded in the $\mathcal{W}+$ space.}
    \label{fig:real_images}
    \vspace{-0.3cm}
\end{figure}

In this section, we demonstrate that the obtained Neural ODE-based edits can be applied to real images projected to the StyleGAN 2 $\mathcal{W}+$ space. We used the standard projector \cite{karras2020analyzing} and trained model of depth $1$. As commonly done for real image editing \cite{abdal2019image2stylegan,abdal2020image2stylegan++,abdal2020styleflow} we apply edits to a subset of indices of $\mathcal{W}+$. Concretely, we used the indices (0-6) for this experiment. Our results are provided at \Cref{fig:real_images}.

\subsubsection{Human evaluation}
Similar to previous works \cite{abdal2020styleflow}, we perform a human evaluation of the quality of the obtained edits. We selected $13$ common attributes for \emph{FFHQ} and $32$ attributes for \emph{Places365}. During the evaluation, we have presented three images to an assessor: an original image and two modified images obtained by two different methods; these two images were shown in random order. We asked the following questions: (\textbf{Q1})`Which has better attribute change to target \texttt{<attr>}?' and (\textbf{Q2})`Which better preserved identity of the original image?'. The possible answers included \textit{Left}, \textit{Right} and \textit{None / both / not applicable}; the total number of participants were $21$ and the number of responses was $\sim 1000$ for both datasets. We compare $\mathbf{Ours(nonlinear)}$ against $\mathbf{Ours(linear)}$ and $\mathbf{IF}$ methods in separate studies. Results are given in \Cref{tab:result_human_eval}; we observe that our nonlinear method allows for better control and identity preservation, especially on the more challenging \emph{Places 365} dataset. We noticed that on this dataset $\mathbf{IF}$ often struggles to make any visual edits, which explains its superiority for $\textbf{Q2}$. We visualize the interface of our questionnaire on \Cref{fig:questions} in \Cref{app:eval}. The breakdown of the improvements for each particular attribute (for the $\mathbf{Ours(nonlinear)}$ vs $\mathbf{IF}$ evaluation) is given in \Cref{fig:eval} in \Cref{app:eval}; we also noticed that the top 4 most challenging attributes coincided for $\mathbf{IF}$ and $\mathbf{Ours(linear)}$, which indicates the need for nonlinearity for certain attributes (\texttt{wavy\_hair}, \texttt{gray\_hair}).


\begin{table}[t]
\centering
\setlength{\tabcolsep}{4pt}
\begin{tabular}{ccc|cc}
\toprule
 &  \multicolumn{2}{c}{\emph{FFHQ}} & \multicolumn{2}{c}{\emph{Places365}}\\
\cmidrule(lr){2-5}
 \textbf{vs} & $\mathbf{IF}$ & $\mathbf{Ours(linear)}$ & $\mathbf{IF}$ & $\mathbf{Ours(linear)}$ \\
\midrule
\textbf{Q1} & $+34\%$ & $+10\%$ & $+47\%$ & $+48\%$ \\
\textbf{Q2} & $+4\%$ &  $+5\%$ &  $-20\%$& $+31\%$\\
\bottomrule
\end{tabular}
\caption{Improvement of $\mathbf{Ours(nonlinear)}$ against linear methods (in absolute percentage values) according to human evaluators.}
\label{tab:result_human_eval}
\vspace{-0.4cm}
\end{table}

\subsubsection{Analysis of learned Neural ODEs}
\begin{figure}[htb!]
    \centering
\includegraphics[width=\linewidth]{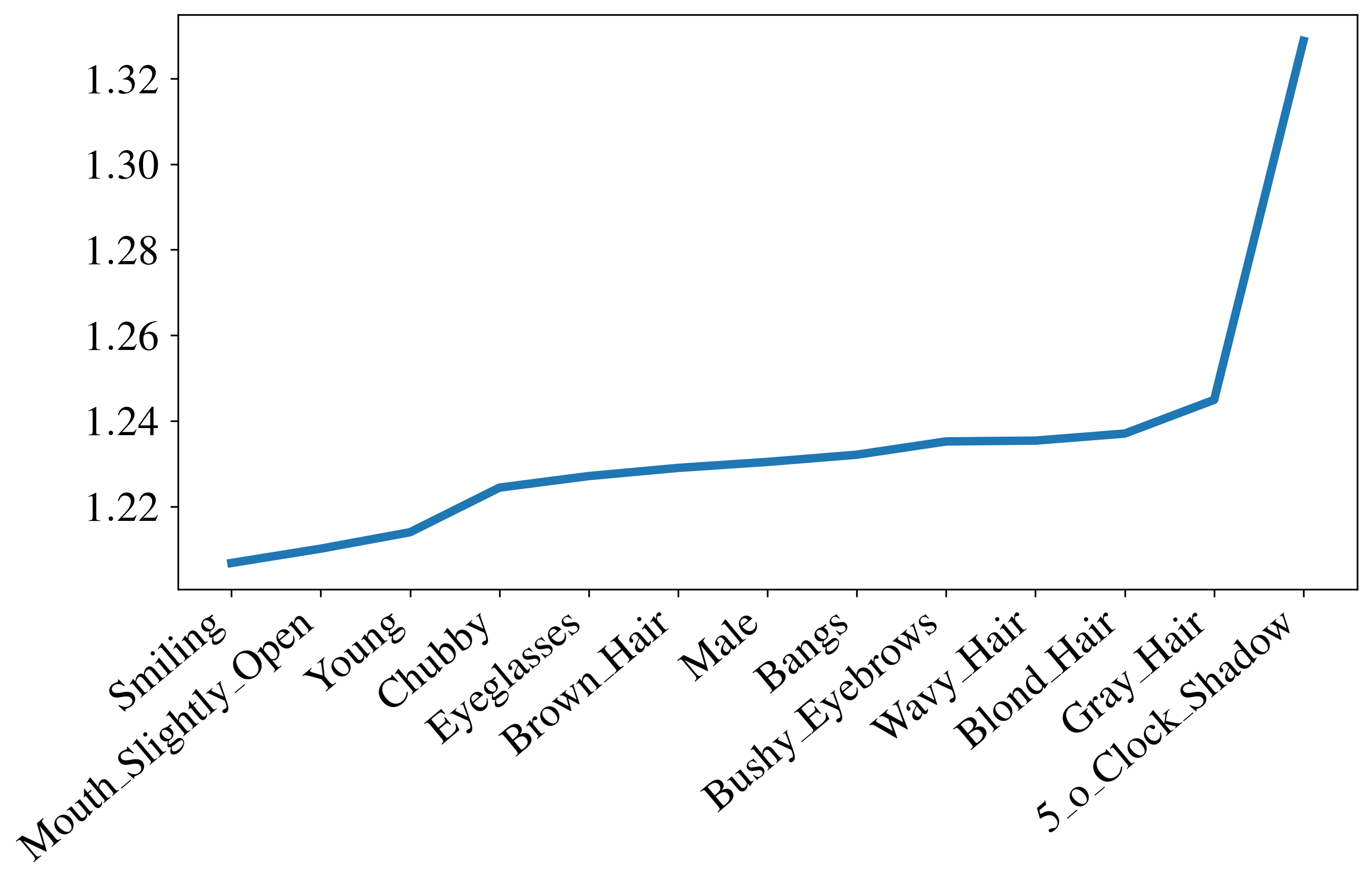}
    \caption{Estimated values of $\mathcal{H}_{SVD}$ for a number of attributes on \emph{FFHQ}. }
    \label{fig:hsvd_ffhq}
\end{figure}

In this section, we study the Neural ODEs obtained for various attributes with our method. We focus on the model of depth $1$, i.e, it takes the form \mbox{$\frac{d\mathbf{w}}{dt}=A\mathbf{w}+b$}. For the analysis, we ignore the normalization of the right-hand side since it does not affect the obtained trajectories and corresponds only to their reparameterization. To study the obtained ODE, it is convenient to switch to the eigenbasis of $A$. In these coordinates (assuming all the eigenvalues are real), the ODE takes the simple form \mbox{$\frac{d\widetilde{\mathbf{w}}}{dt} = \mathrm{diag}(\lambda_1, \hdots \lambda_N) \widetilde{\mathbf{w}} + \widetilde{b}$}. The eigenvalues of large magnitude $| \lambda_i | \gg 1$ correspond to a `fast' subspace where some nontrivial dynamics happens. On the other hand, in the `slow' subspace with $| \lambda_i | \ll 1$, the dynamics is close to linear, i.e., the trajectories are close to straight lines. Thus, we can measure the `complexity' of an attribute by evaluating how quickly the eigenvalues of the corresponding matrix decay. If they decay rapidly, then this attribute is easier to control with linear shifts and requires more `nonlinear' controls in the opposite case. One way to estimate how many vectors span the range of the matrix $A$ is via \emph{singular entropy}. It is defined in terms of singular values $\lbrace \sigma_i \rbrace$ of the matrix $A$ in the following way:
\begin{equation}
    \mathcal{H}_{SVD}(A) = -\sum_{i=1}^N \tilde{\sigma}_i \log \tilde{\sigma}_i, 
\end{equation}
with $\lbrace \tilde{\sigma}_i \rbrace$ being the set of normalized singular values: $\tilde{\sigma}_i = \sfrac{\sigma_i}{\sum \sigma_i}$. The values of $\mathcal{H}_{SVD}(A)$ can serve as a proxy to the log-dimensionality of the `fast' subspace of $A$. We hypothesize that for the attributes with larger values of $\mathcal{H}_{SVD}$, our nonlinear method provides a more significant improvement. When computing $\mathcal{H}_{SVD}$, we utilize the first $128$ singular values (out of $512$) in order to get rid of the noise induced by the training procedure (the results are not sensitive to this parameter). The obtained values are provided at \Cref{fig:hsvd_ffhq}. 
To verify our hypothesis, we compute the Spearman rank correlation between the attribute ordering provided by $\mathcal{H}_{SVD}$ and human evaluation ordering visualized at \Cref{fig:eval}. The obtained value is $\sim 0.41$, confirming the existence of a correlation. Interestingly, we find that even such `simple' attributes such as \texttt{gray\_hair} still require a nontrivial trajectory in the latent space. The obtained $\mathcal{H}_{SVD}$ values for \emph{Places365} are available in the supplementary material. Overall, based on the experimental results, we argue that attributes requiring a `global' content change can not be adequately controlled with linear edits. E.g., for \texttt{gray\_hair} which is naturally entangled with the facial appearance, we do not simply change the hair color but also make the entire face older. Similar logic holds, for instance, for the \texttt{lush} attribute on \emph{Places365}. On the other hand, such attributes as \texttt{Smiling} or \texttt{Bushy\_Eyebrows} require relatively small and localized changes, and we observe that the IF method is on par with our nonlinear model.

\section{Conclusion}
In this work, we discussed a novel approach for image manipulations via nonlinear shifts, parameterized by a Neural ODE model. On multiple datasets, we demonstrated an advantage of our approach over standard linear shifts. For the analysis, we simply considered state-of-the-art StyleGAN 2 trained in a conventional manner. Thus, it may be possible that design choices for this model do not allow for achieving perfect disentanglement. One interesting direction for future work is to better understand the arrangement of attribute distributions in the latent space and how it can be utilized to achieve better disentanglement. Another possible direction to achieve this goal would be to try to tune the GAN architecture so it better incorporates geometrical (\ie, shape) inductive biases.

\clearpage
{\small
\bibliographystyle{ieee_fullname}
\bibliography{egbib}
}

\clearpage
\newpage
\appendix
\onecolumn
\section{Model details}{\label{app:archi}}
\subsection{Code and pretrained models}
Our code and models used for experiments are available at this url: \href{https://github.com/KhrulkovV/nonlinear-image-editing}{github}.
\subsection{GANs}
In our experiments we used \url{https://github.com/rosinality/stylegan2-pytorch} for training GANs for \emph{MPI3D} and \emph{Isaac3D}. See the list of training parameters below.
\begin{table*}[htb!]
\parbox{\linewidth}{
    \vspace{0.3em}
    \centering
    \begin{tabular}{l|l}
    \toprule
    {\textbf{Parameter}} & {\textbf{Value}} \\
    \midrule
    \texttt{iter} & $800000$ for \emph{Isaac3D} and $300000$ for \emph{MPI3D} \\
    \texttt{batch} & $32$ \\
    \texttt{n\_sample} & $64$ \\
    \texttt{size} & $128$ for \emph{Isaac3D} and $64$ for other \emph{MPI3D}\\
    \texttt{r1} & $10$ \\
    \texttt{path\_regularize} & $2$ \\
    \texttt{path\_batch\_shrink} & $2$ \\
    \texttt{d\_reg\_every} & $16$ \\
    \texttt{g\_reg\_every} & $4$ \\
    \texttt{mixing} & $0.9$ \\
    \texttt{lr} & $0.002$ \\
    \texttt{augment} & False \\
    \texttt{augment\_p} & $0$ \\
    \texttt{ada\_target} & $0.6$ \\
    \texttt{ada\_length} & $500000$ \\
    \texttt{latent\_dim} & $512$ \\
    \texttt{n\_mlp} & $3$  \\
     \texttt{width} & $512$ \\
    \midrule
    \texttt{truncation} & $1.0$ for \emph{Isaac3D} and $0.7$ for \emph{MPI3D} \\ 
    \texttt{mean\_latent} & $4096$ \\
    \texttt{input\_is\_latent} & \texttt{True} \\
    \texttt{randomize\_noize} & \texttt{False} (for evaluation) \\
    \bottomrule
    \end{tabular}
}
\label{tab:hyperparams_stylegan2}
\end{table*}
\\
For \emph{FFHQ} we took the pretrained checkpoint \texttt{ffhq-res256-mirror-paper256-noaug} provided at \url{https://github.com/NVlabs/stylegan2-ada}. For \emph{CUB-200-2011} and \emph{Places365} we trained StyleGAN 2 with ADA using the config \texttt{auto} from the above repository. We only changed the number of layers in the style network to $2$ for the \emph{CUB-200-2011} dataset and to $8$ for \emph{Places365}. All the \texttt{tensorflow} checkpoints were converted to the \texttt{Pytorch} format for the analysis. For these models we used truncation $0.7$ when generating samples.
\subsection{Attribute regressors}
All our attribute regressors have the same structure: a convolutional backbone followed by a multiclass classification head (represented as an MLP of depth $2$). We used the following backbones:
\begin{itemize}
    \item A simple 4-layer CNN for \emph{MPI3D},
    \item Randomly initialized ResNet18 for \emph{Isaac3D},
    \item ResNet18 pretrained on ImageNet for \emph{FFHQ}, \emph{CUB-200-2011}, \emph{Places365}.
\end{itemize}
\clearpage
\newpage
\subsection{Disentanglement via pretrained encoders}
As described in the main text, we consider an alternative loss formulation to enforce disentanglement. Namely, we take the pretrained FaceNet $\mathcal{F}$ and utilize 
$$
\mathcal{L}_2(\mathbf{w}; \theta) = -\mathrm{cos}
\Big(\mathcal{F}[G(\mathbf{w}(T, \theta)], \mathcal{F}[G(\mathbf{w})]\Big),
$$
i.e., we simply want to keep the identity of a person after editing.
The final loss function takes the form $\mathcal{L} = \mathcal{L}_1 + \alpha \mathcal{L}_2$ where $\alpha=0.5$ was chosen (we found it not to significantly affect the results). We trained the model using exactly the same experimental setup for the same $13$ attributes as in the human evaluation study in \Cref{sec:experiments}. We utilized the InceptionResnetV1 model pretrained on the VGGFace2 \cite{cao2018vggface2} dataset available at \href{https://github.com/timesler/facenet-pytorch}{https://github.com/timesler/facenet-pytorch}. To evaluate the results, we performed human evaluation asking ``Which edited image better preserved identity of the original image?''. Interestingly, we found that the FaceNet disentanglement approach provided better results ($60\%$ vs $40\%$), while correctly performing the edits.

\clearpage
\newpage

\section{Additional examples}{\label{app:examples}}
\begin{figure*}[htb!]
    \centering
    \includegraphics[width=0.9\linewidth]{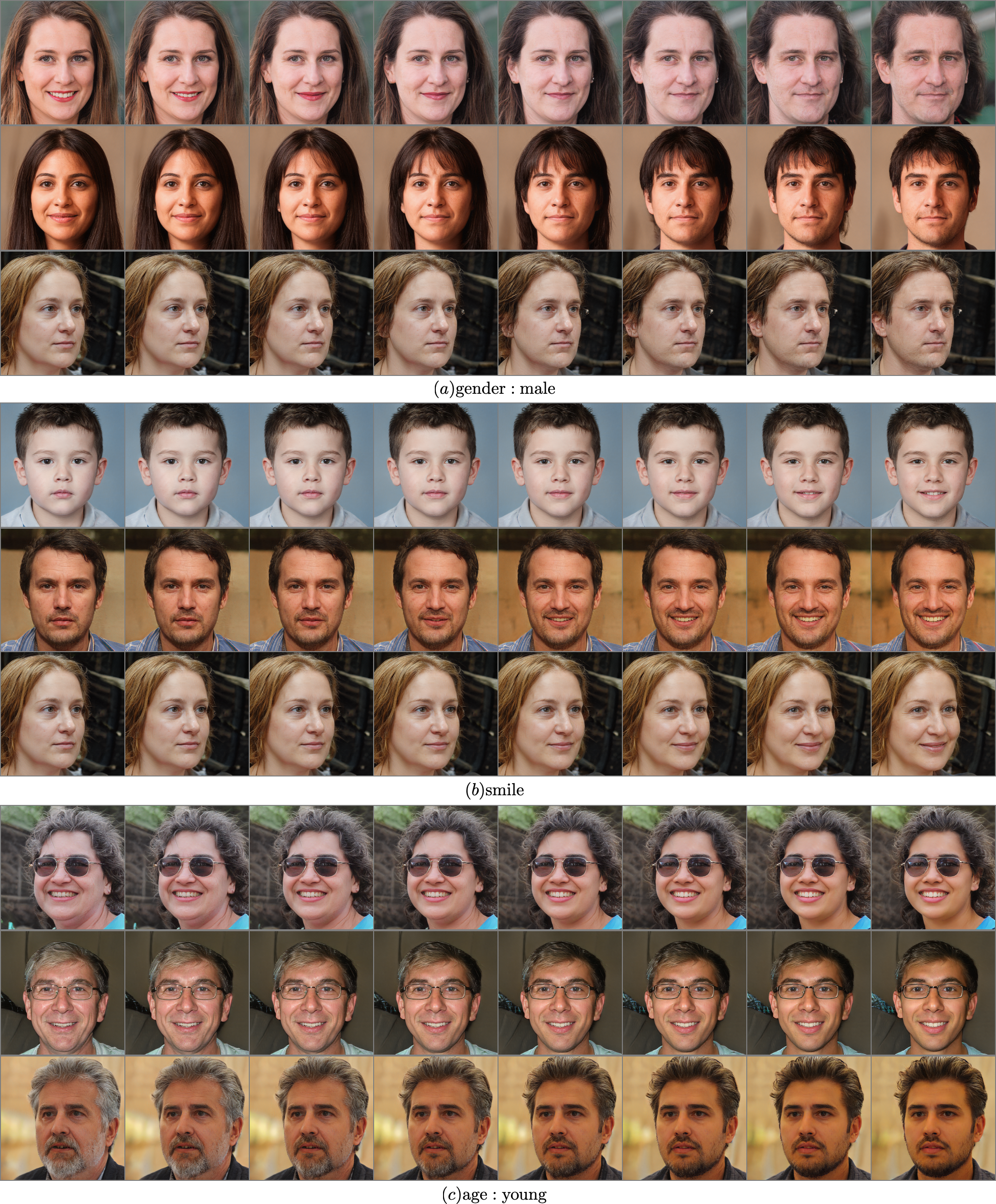}
    \caption{Additional examples of factor manipulations on \emph{FFHQ} made by our Neural ODE method (nonlinear).}
    \label{fig:celeba_many}
    \vspace{-0.3cm}
\end{figure*}
\begin{figure*}[htb!]
    \centering
    \includegraphics[width=0.9\linewidth]{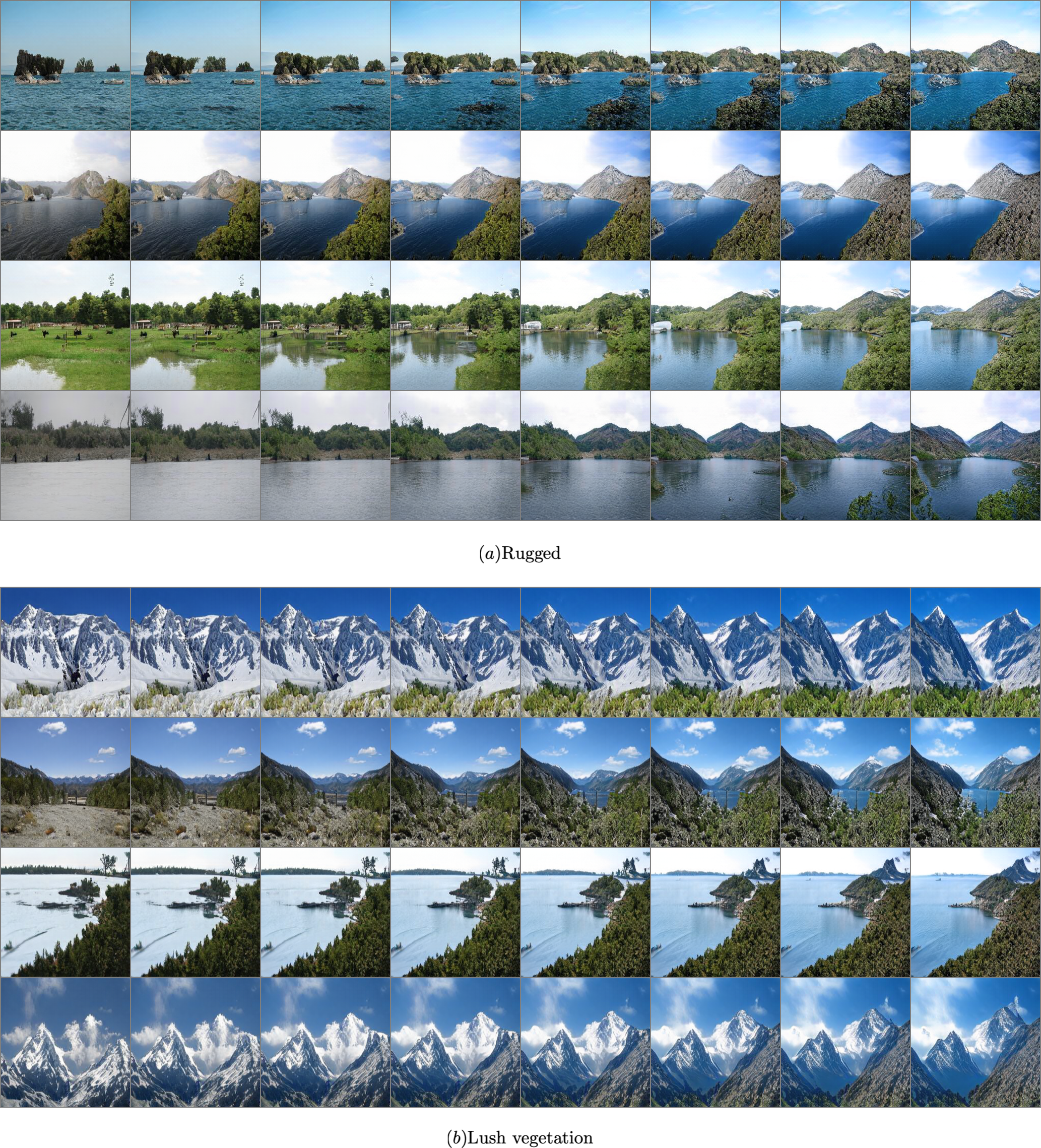}
    \caption{Additional examples of factor manipulations on \emph{Places365} made by our Neural ODE method (nonlinear).}
    \label{fig:celeba_many}
\end{figure*}
\clearpage
\newpage 

\subsection{CUB-200-2011}{\label{app:cub}}
\emph{CUB-200-2011} consists of $11,788$ images of birds with given attribute information covering $312$ binary factors of variation \cite{WahCUB_200_2011}. These attributes represent visual features focusing mostly on color, patterns, shapes, \etc. For \emph{CUB-200-2011} we utilized the existent annotations. We selected the attributes with at least $10\%$ of positive examples for both values, resulting in $109$ attributes total. Results for this dataset are available in the supplementary material.

\begin{figure}[htb!]
    \centering
\includegraphics[width=0.6\linewidth]{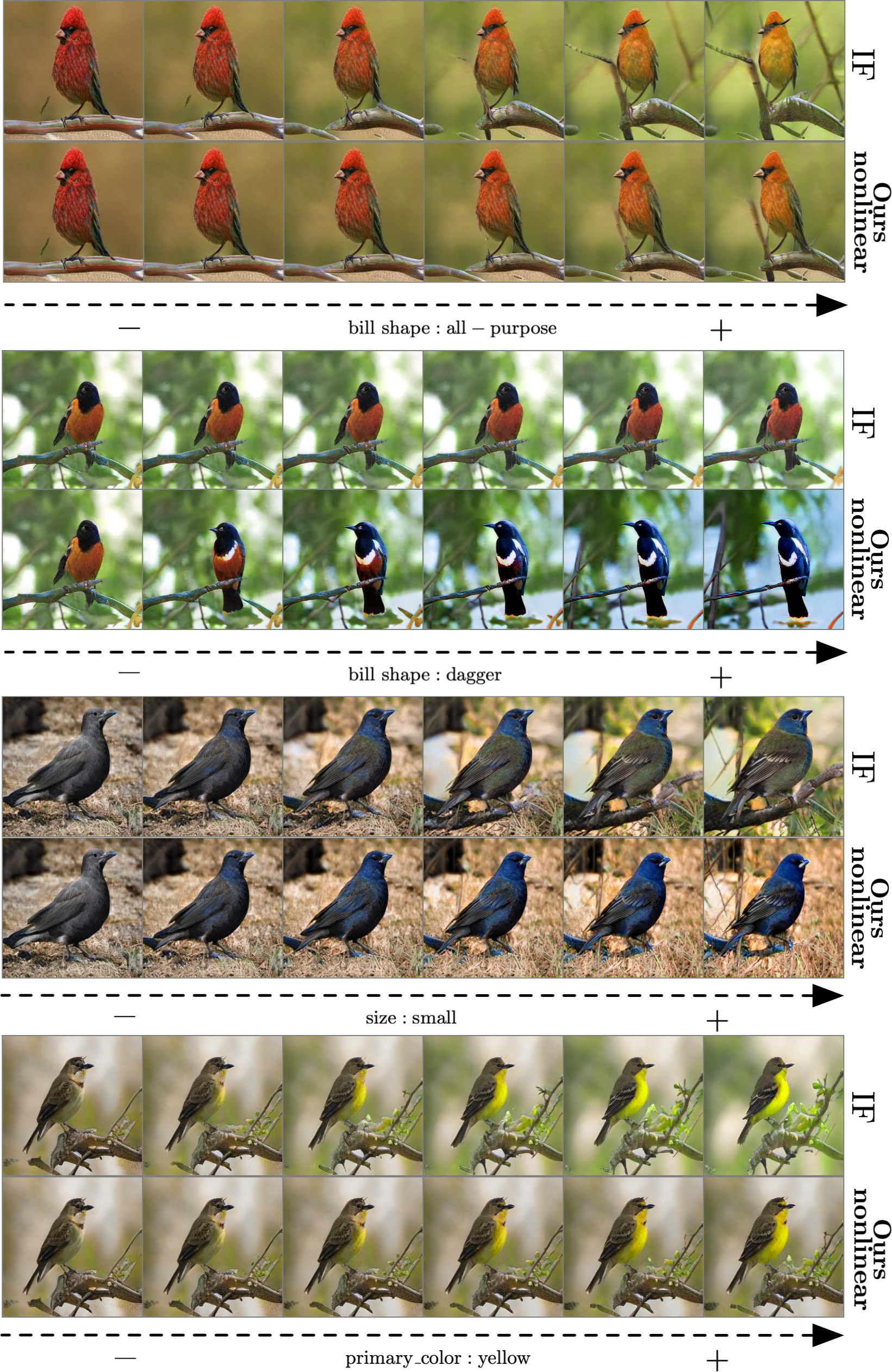}
    \caption{Manipulating attributes corresponding to the birds bill shape, size, and color on the \emph{CUB-200-2011} dataset. While in both cases, we observe some degree of entanglement, linear shifts produced by $\mathrm{IF}$ do not change the attribute of choice (\eg, for the first three attributes) or change the obtained factor in an unnatural manner (manipulated \texttt{primary color:yellow} looks slightly greenish).}
    \label{fig:birds}
\end{figure}

This dataset was particularly challenging to experiment with since the main bulk of its attributes was connected to texture-based features. In our experimental setup, we chose the following factors: the bill shape, the bird size, and one texture attribute describing the primary color; our findings are summarized in \Cref{fig:birds}. We may see a significant difference in bill shape and body size attributes and less visually perceptive difference for feather color. However, we may notice that in several cases, both methods produce entangled modification, \eg, changing the bird color simultaneously with size manipulation. 
\clearpage
\newpage 

\section{Human evaluation}{\label{app:eval}}
\begin{figure*}[htb!]
    \centering
    \includegraphics[width=0.9\linewidth]{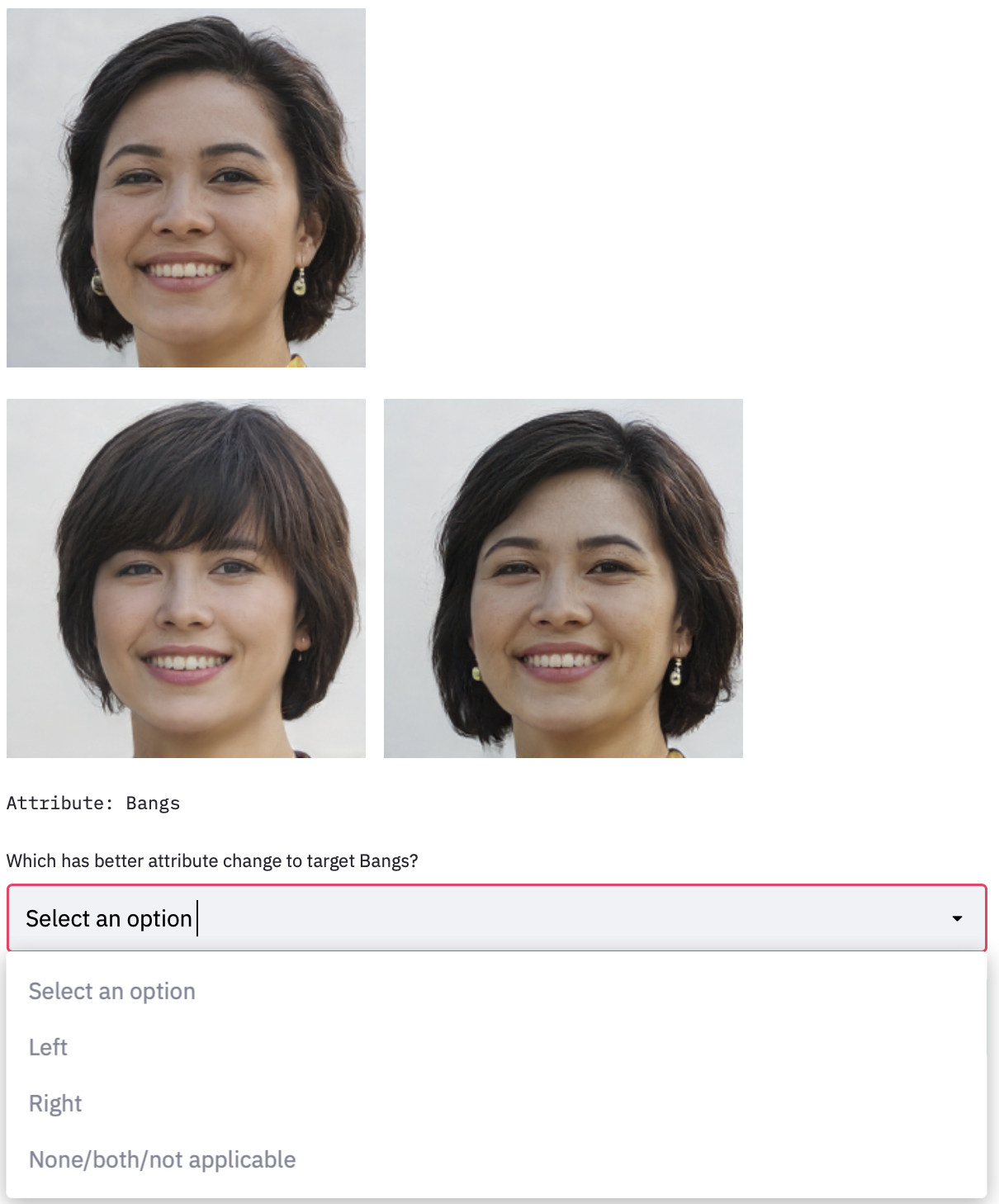}
    \caption{The interface of human evaluation questionnaire.}
    \label{fig:questions}
\end{figure*}
\clearpage
\newpage 
\subsection{Attribute breakdown}
\begin{figure*}[htb!]
    \centering
\includegraphics[width=0.6\linewidth]{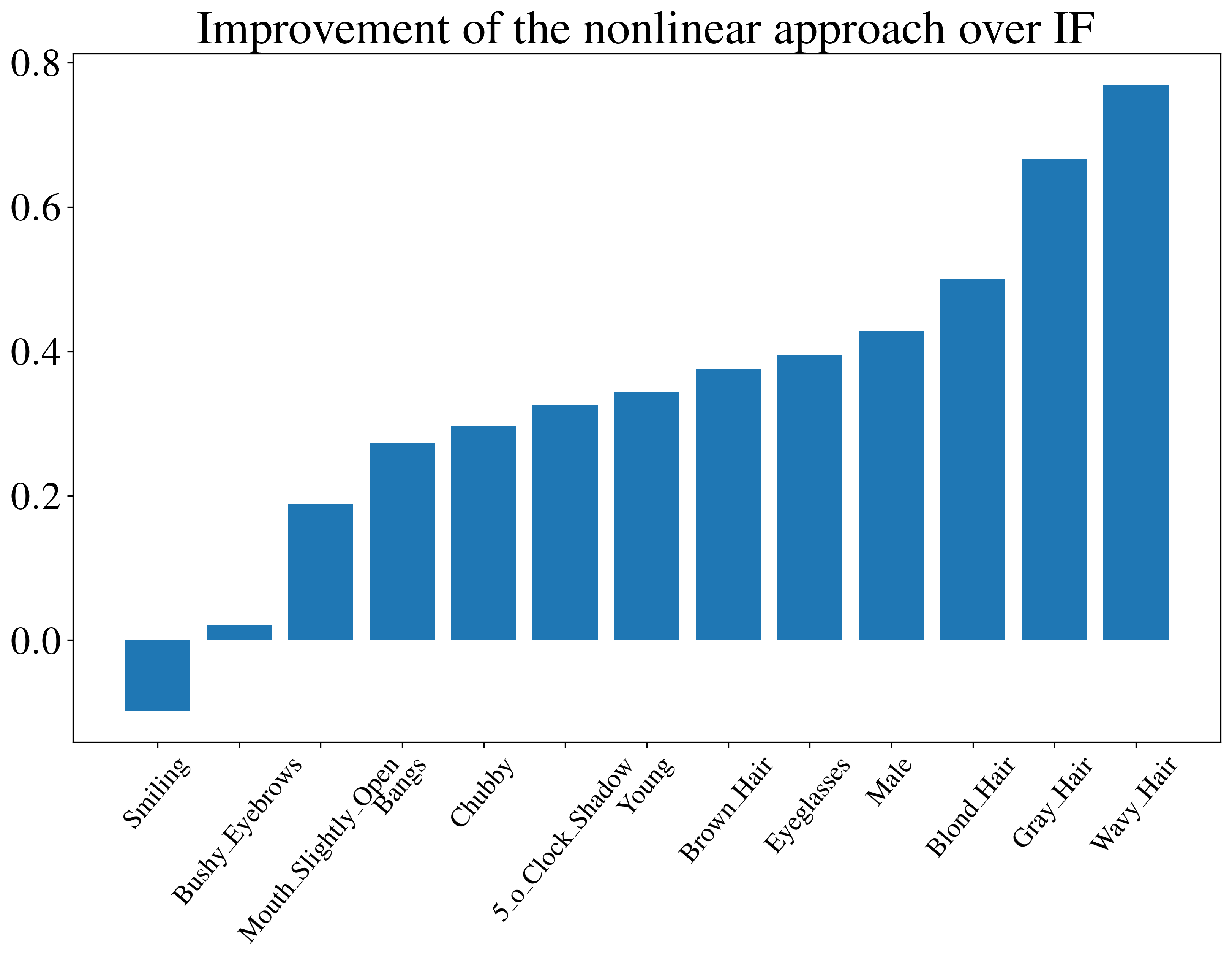}
    \caption{The breakdown of the improvement of the nonlinear model over IF on the \emph{FFHQ} dataset.}
    \label{fig:eval}
\end{figure*}

\clearpage
\newpage

\section{$\mathcal{H}_{SVD}$ for \emph{Places365}}
Here we provide the obtained values of $\mathcal{H}_{SVD}$ for all the attributes on the \emph{Places365} dataset.
\begin{figure*}[htb!]
    \centering
    \includegraphics[width=0.9\linewidth]{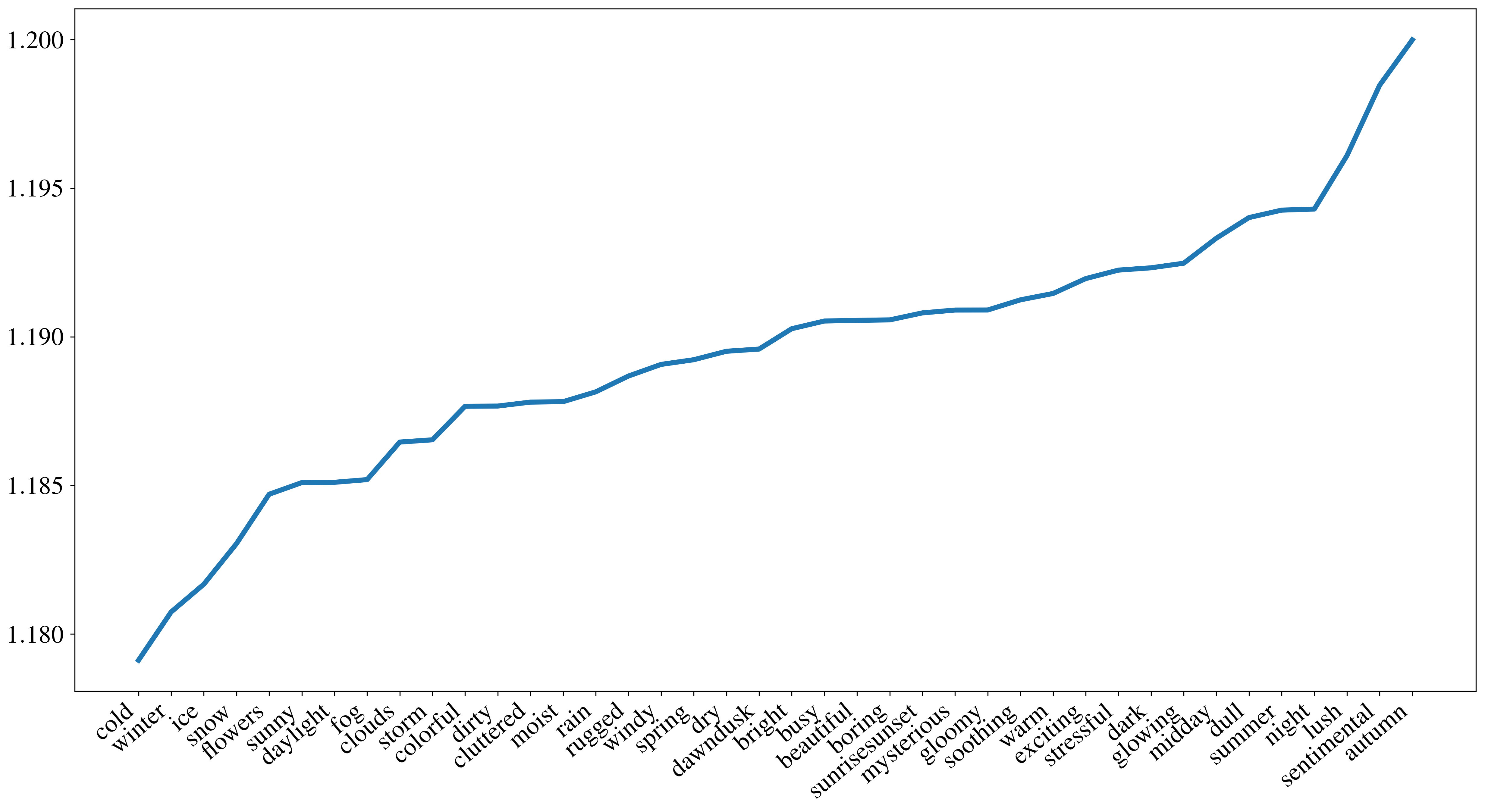}
    \caption{The values of $\mathcal{H}_{SVD}$ for various attributes on \emph{Places365}}
    \label{fig:scenes_att}
\end{figure*}
For this dataset, we do not compute the correlation between the human evaluation breakdown and entropy scores since, in most of the cases, our nonlinear method significantly outperformed IF.
We note that the easiest attributes according to \Cref{fig:scenes_att}, such as \texttt{cold}, \texttt{winter}, \texttt{ice}, correspond to simple color based transformations. On the other hand, the most difficult ones such \texttt{lush} are mostly `content' based.

\end{document}